\newif\ifarxiv
\arxivtrue
\documentclass[12pt]{article}
\usepackage{amsfonts}
\usepackage{amssymb}
\usepackage{graphicx,float}
\usepackage{color}
\usepackage{amsmath}
\usepackage{amsthm}
\usepackage{setspace}
\usepackage{fullpage}
\usepackage{enumitem}
\usepackage[colorlinks,citecolor=blue,urlcolor=blue,filecolor=blue,backref=page]{hyperref}
\usepackage{setspace}
\newcommand{\defeq}{\mathrel{\mathop:}=}
\setenumerate{label=(\alph*),noitemsep, topsep=5pt}

\graphicspath{{./fig/}}

\begin{document}
\renewcommand*{\thefootnote}{\fnsymbol{footnote}}

\title{\vspace{-1cm} Deep Learning }

\author{Nicholas G. Polson \footnote {Booth School of Business,  
University of Chicago.} \and Vadim O. Sokolov  \footnote{George Mason University.}}

\date{\small{First Draft: December 2017\\
		This Draft: July 2018
		}}

\maketitle


\begin{abstract}
\bigskip

\noindent Deep learning (DL) is a high dimensional data reduction technique for constructing high-dimensional predictors in input-output models. DL is a form of machine learning that uses hierarchical layers of latent features. In this article, we review the state-of-the-art of deep learning from a modeling and algorithmic perspective. We provide a list of successful areas of applications in Artificial Intelligence (AI), Image Processing, Robotics and Automation. Deep learning is predictive in its nature rather then inferential and can be viewed as a black-box methodology for high-dimensional function estimation.\\

\end{abstract} \vspace{-.6cm}

\newpage
\singlespacing

\section{Introduction}
Prediction problems are of great practical and theoretical interest. Deep learning is a form of machine learning which provides a tool box for high-dimensional function estimations. It uses hierarchical layers of hidden features to model complex nonlinear high-dimensional input-output models. As statistical predictors, DL have a number of  advantages over traditional approaches, including
\begin{enumerate}
\item input data can include all data of possible relevance to the prediction problem at hand
\item nonlinearities and complex interactions among input data are accounted for seamlessly
\item overfitting is more easily avoided than traditional high dimensional procedures
\item there exists fast, scale computational frameworks (\verb|TensorFlow|, \verb|PyTorch|).
\end{enumerate}

There are many successful applications of deep learning across many fields, including speech recognition, translation, image processing, robotics, engineering, data science, healthcare among many others. These applications include algorithms such as

\begin{enumerate}
	\item Google Neural Machine Translation \cite{wu2016google} closes the gap with humans in accuracy of the translation by 55-85\% (estimated by people on a 6-point scale). One of the keys to success of the model is the use of Google's huge dataset.
	\item Chat bots which predict natural language response have been available for many years. Deep learning networks can significantly improve the performance of chatbots \cite{henderson2017efficient}. Nowadays they provide help systems for companies and home assistants such as Amazon's Alexa and Google home.
	\item Google WaveNet (developed by DeepMind \cite{oord2016wavenet}), generates speech from text and  reduces the gap between the state of the art and human-level performance by over 50\% for both US English and Mandarin Chinese.
	\item Google Maps were improved after deep learning was developed to analyze more then 80 billion Street View images and to extract names of roads and businesses \cite{wojna2017attention}.
	\item Health care diagnostics were developed using Adversarial Auto-encoder model found new molecules to fight cancer.  Identification and generation of new compounds was based on available biochemical data \cite{kadurin2017cornucopia}.
	\item Convolutional Neural Nets (CNNs), which are central to image processing, were developed to detect pneumonia from chest X-rays with better accuracy then practicing radiologists \cite{rajpurkar2017chexnet}. Another CNN model is capable of identifying skin cancer from biopsy-labeled test images \cite{esteva2017dermatologist}.
	\item \cite{shallue2017identifying} discovered two new planets using deep learning and data from  NASA's Kepler Space Telescope.
	\item In more traditional engineering, science applications, such as spatio-temporal and financial analysis deep learning showed superior performance compared to traditional statistical learning techniques \cite{polson_deep_2017,dixon2017deep,heaton2017deep,sokolov2017discussion,feng2018deep,feng2018deepa}
\end{enumerate}
The rest of our article proceeds as follows. Section \ref{sec:dl} reviews mathematical aspects of deep learning and popular network architectures. Section \ref{sec:algo} provides overview of algorithms used for estimation and prediction. Finally, Section \ref{sec:gp} discusses some theoretical results related to deep learning. 

\section{Deep Learning}\label{sec:dl}
Deep learning is data intensive and provides predictor rules for new high-dimensional input data.  The fundamental problem is to find a predictor $ \hat{Y} (X) $ of an output $Y$. Deep learning trains a model on data by passing learned features of data through different ``layers" of hidden features.  That is, raw data is entered at the bottom level, and the desired output is produced at the top level, the result of learning through many levels of transformed data. Deep learning is hierarchical in the sense that in every layer, the algorithm extracts features into factors, and a deeper level's factors become the next level's features. 

Consider a high dimensional matrix $X$ containing a large set of potentially relevant data.   Let $Y$ represent an output (or response) to a task which we aim to solve based on the information in $X$. This leads to an input-output map $Y = F(X)$ where $X=(X_1,\ldots,X_p)$.
\cite{breiman_statistical_2001} summaries the difference between statistical and machine learning philosophy as follows.
\begin{quote}
\small{\emph{
``There are two cultures in the use of statistical modeling to reach conclusions from data. One assumes that the data are generated by a given stochastic data model. The other uses algorithmic models and treats the data mechanism as unknown.}}
\end{quote}\vspace{-0.8cm}

\begin{quote}
\small{\emph{
Algorithmic modeling, both in theory and practice, has developed rapidly in fields outside statistics. It can be used both on large complex data sets and as a more accurate and informative alternative to data modeling on smaller data sets. If our goal as a field is to use data to solve problems, then we need to move away from exclusive dependence on data models and adopt a more diverse set of tools.''}}
\end{quote}

\subsection{Network Architectures}
A deep learning architecture can be described as follows. Let $ f_1 , \ldots , f_L $ be given \emph{univariate} activation functions for each of the $L$ layers. Activation functions are nonlinear transformations of weighted data. A semi-affine activation rule is then defined by
$$
f_l^{W,b} \defeq f_l \left ( \sum_{j=1}^{N_l} W_{lj} X_j + b_l \right ) = f_l ( W_l X_l + b_l )\,,
$$
which implicitly needs the specification of the number of hidden units $N_l$. Our deep predictor, given the number of layers $L$, then becomes the composite map
\begin{equation}
\hat{Y}(X) \defeq F(X) = \left ( f_l^{W_1,b_1} \circ \ldots \circ f_L^{W_L,b_L} \right ) ( X)\,.\label{DLComp}
\end{equation}

The fact that DL forms a universal `basis' which we recognise in this formulation dates to Poincare and Hilbert is central. From a practical perspective, given a large enough data set of ``test cases",  we can empirically learn an optimal predictor.

Similar to a classic basis decomposition, the deep approach uses univariate activation functions to decompose a high dimensional $X$.

Let $ Z^{(l)} $ denote the $l$th layer, and so $ X = Z^{(0)}$.
The final output $Y$ can be numeric or categorical.
The explicit structure of a deep prediction rule is then
\begin{align*}
\hat{Y} (X) & = W^{(L)} Z^{(L)} + b^{(L)} \\
Z^{(1)} & = f^{(1)} \left ( W^{(0)} X + b^{(0)} \right ) \\
Z^{(2)} & = f^{(2)} \left ( W^{(1)} Z^{(1)} + b^{(1)} \right ) \\
\ldots  & \\
Z^{(L)} & = f^{(L)} \left ( W^{(L-1)} Z^{(L-1)} + b^{(L-1)} \right )\,.
\end{align*}
Here $ W^{(l)} $ is a weight matrix and $ b^{(l)} $ are threshold or activation levels.
Designing a good predictor depends crucially on the choice of univariate activation functions $ f^{(l)} $.
The $Z^{(l)}$ are hidden features which the algorithm will extract. 

Put differently, the deep approach employs hierarchical predictors comprising of a series of $L$ nonlinear transformations applied to $X$. Each of the $L$ transformations is referred to as a \emph{layer}, where the original input is $X$,
the output of the first transformation is the first layer, and so on, with the output $\hat{Y} $ as the first layer. The layers $1$ to $L$ are called \emph{hidden layers}. The number of layers $L$ represents the \emph{depth} of our routine.

\begin{figure}[H]
	\begin{tabular}{ccc}
		\includegraphics[width=0.18\textwidth]{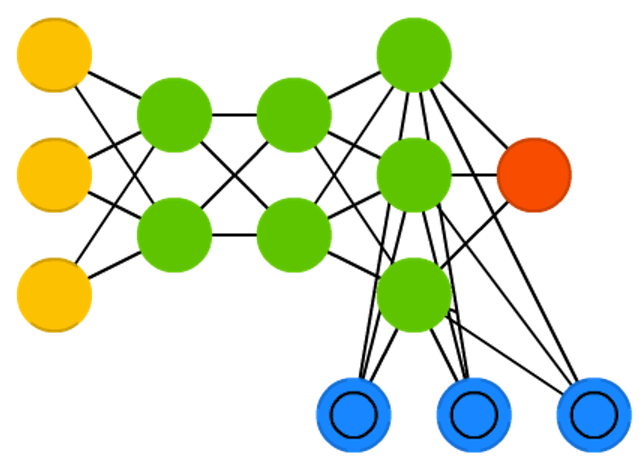} & \includegraphics[width=0.18\textwidth]{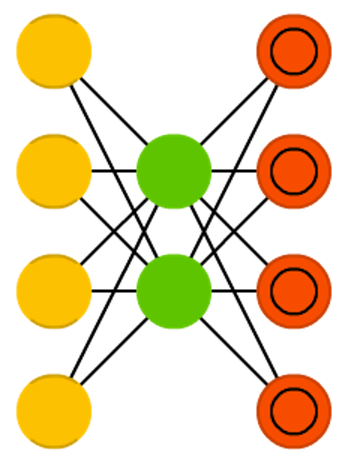} & \includegraphics[width=0.28\textwidth]{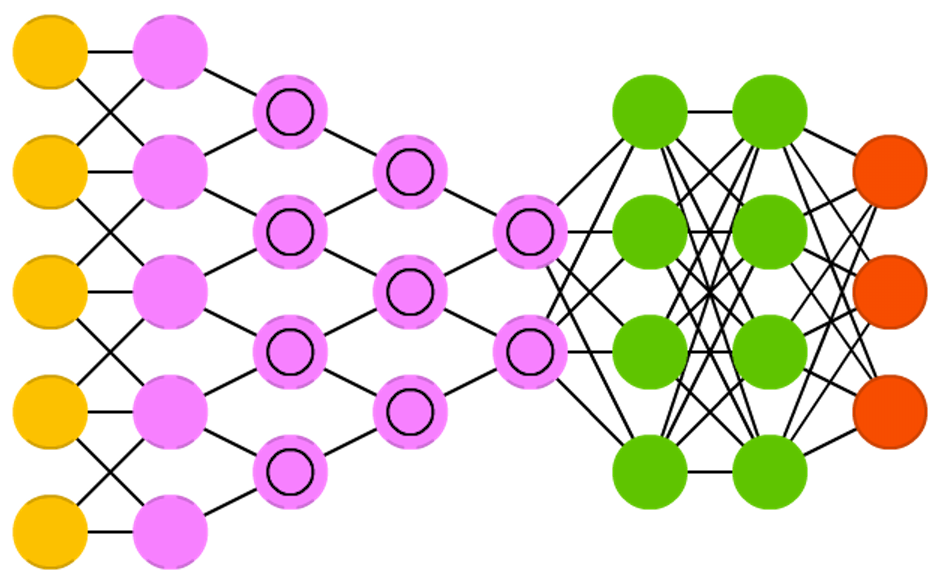}\\ 
		Neural Turing Machine & Auto-encoder & Convolution\\
		\includegraphics[width=0.23\textwidth]{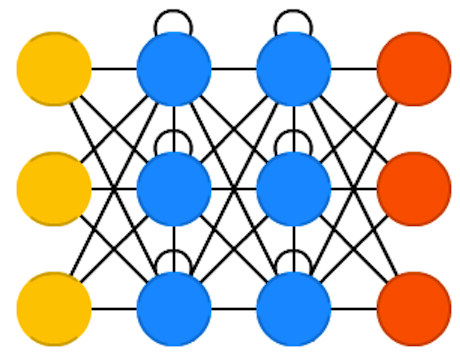} & \includegraphics[width=0.23\textwidth]{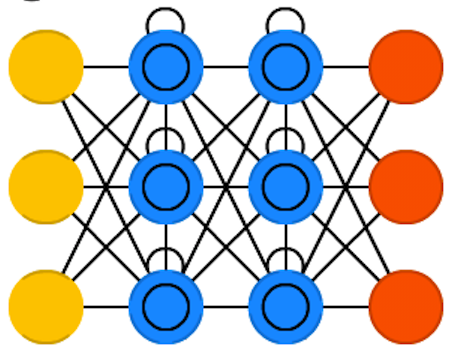}&
		\includegraphics[width=0.3\textwidth]{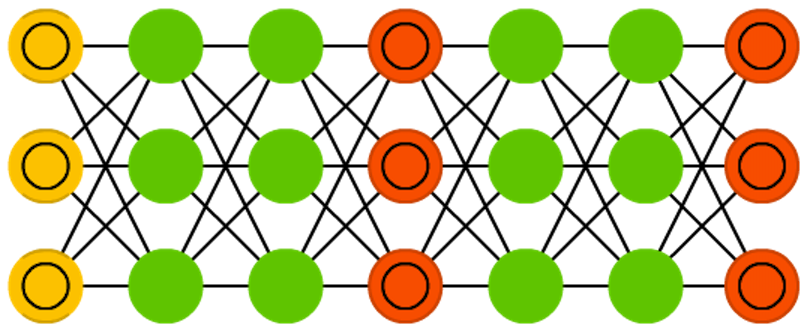}\\
		Recurrent & Long / short term memory & GAN
	\end{tabular}
	\caption{Commonly used deep learning architectures. Each circle is a neuron which calculates a weighted sum of an input vector plus bias and applies a non-linear function to produce an output. Yellow and red colored neurons are input-output cells correspondingly. Pink colored neurons apply weights inputs using a kernel matrix. Green neurons are hidden ones. Blue neurons are recurrent ones and they append its values from previous pass to the input vector. Blue neuron with circle inside a neuron corresponds to a memory cell. Source:	\url{http://www.asimovinstitute.org/neural-network-zoo}.}
	\label{fig:arch}
\end{figure}

Figure \ref{fig:arch} illustrates a number of commonly used structures; for example, feed-forward architectures, auto-encoders, convolutional, and neural Turing machines. Once you have learned the dimensionality of the weight matrices which are non-zero, there's an implied network structure. 

\paragraph{Stacked GLM.} From a statistical viewpoint, deep learning models can be viewed as stacked Generalized Linear Models \cite{polson2017}. The expectation over dependent variable in GLM is computed using affine transformation (linear regression model) followed by a non-linear univariate function (inverse of the link function). GLM is given by
\[
E(y \mid x) = g^{-1}(w^Tx).
\]
Choice of link function is defined by the target distribution $p(y \mid x)$. For example when $p$ is binomial, we choose $g^{-1}$ to be sigmoid $1/(1+\exp(-w^Tx))$.

Recently deep architectures (indicating non-zero weights) include convolutional neural networks (CNN), recurrent NN (RNN), long short-term memory (LSTM), and neural Turing machines (NTM).
\cite{pascanu_how_2013} and \cite{montufar_when_2015} provide results on the advantage of representing some functions compactly with deep layers. 
\cite{poggio_deep_2016} extends theoretical results on when deep learning can be exponentially better than shallow learning. 
\cite{bryant_analysis_2008} implements \cite{sprecher_survey_1972} algorithm to estimate the non-smooth inner link function.  In practice, deep layers allow for smooth activation functions to provide ``learned''  hyper-planes which find the underlying complex interactions and regions  without having to see an exponentially large number of training samples.

Commonly used activation functions are sigmoidal ($\cosh$ or $\tanh$), heaviside gate functions $ I(x > 0 )$, or  rectified linear units (ReLU) $\max\{\cdot,0\}$.  ReLU's especially have been found~\cite{schmidt2017nonparametric} to lend themselves well to rapid dimension reduction. A deep learning predictor is a data reduction scheme that avoids the curse of dimensionality through the
use of univariate activation functions. One particular feature is that the weight matrices  $W_l  \in \Re^{N_l \times N_{l-1}} $ are matrix valued.
This gives the predictor great flexibility to uncover nonlinear features of the data -- particularly so in finance data as the estimated hidden features
$ Z^{(l)} $ can be given the interpretation of portfolios of payouts. The choice of the dimension $ N_l $ is key, however, since
if a hidden unit (aka columns of $W_l$) is dropped  at layer $l$ it kills all terms above it in the layered hierarchy.

\subsection{Autoencoder}

An autoencoder is a deep learning routine which trains $F(X)$ to approximate $X$ (i.e., $X=Y$) via a \emph{bottleneck} structure, which means we select a model $F= f_l^{W_1,b_1} \circ \ldots \circ f_L^{W_L,b_L}$ which aims to concentrate the information required to recreate $X$. Put differently, an autencoder creates a more cost effective representation of $X$.

For example, for a static autoencoder with two linear layers (a .k.a. traditional factor model), we write
$$
Y = W_1 \left ( W_2 X \right )\,.
$$
where $ x \in \Re^k $ and $ W_{hidden} \in \Re^{ p \times k} $ and $ W_{out} \in \Re^{ k \times p } $ where $ p \ll k $.
The goal of an autoencoder is to train the weights so that $ y= x$ with loss function typically given by squared errors. 

If $ W_2$ is estimated from the structure of the training data matrix, then we have a traditional factor model, and the
$W_1$ matrix provides the factor loadings. We note, that principal component analysis (PCA) in particular falls into this category, as we have seen in \eqref{PCA_eq}. If $W_2$ is estimated based on the pair $\hat{X}=\{Y,X\}=X$ (which means estimation of $W_2$ based on the structure of the training data matrix with the specific autoencoder objective), then we have a sliced inverse regression model. If $W_1$ and $W_2$ are simultaneously estimated based on the training data $X$, then we have a two layer deep learning model. 

A dynamic one layer autoencoder for a financial time series $(Y_t)$ can, for example, be written as a coupled system of the form
$$
Y_t = W_x X _t + W_y Y_{t-1} \; \; {\rm and} \; \; \left ( \begin{array}{c}
X_t\\
Y_{t-1} 
\end{array}
\right ) = W Y_t\,.
$$
We then need to learn the weight matrices $W_x$ and $W_y$. Here, the state equation encodes and 
the matrix $W$ decodes the $Y_t$ vector into its history $Y_{t-1}$ and the current state $X_t$.

The auto encoder demonstrates nicely that in deep learning we do not have to model the variance-covariance matrix explicitly, as our model is already directly in
predictive form. (Given an estimated nonlinear combination of deep learners, there is an implicit variance-covariance matrix, but that is not the driver of the method.) 

\subsection{Factor Models}
Almost all shallow data reduction techniques can be viewed as consisting of a low dimensional auxiliary
variable $Z$ and a prediction rule specified by a
composition of functions 
\begin{align*}
\hat{Y} &= f_1^{W_1,b_1} (f_2( W_2X +b_2)\big) 
\\
&= f_1^{W_1,b_1}(Z),\,\text{ where $Z:=f_2(W_2X+b_2)$\,. }
\end{align*}
In this formulation, we also recognise the previously introduced deep learning structure \eqref{DLComp}.
The problem of high dimensional data reduction in general is to find the $Z$-variable and to estimate the layer functions $(f_1, f_2)$ correctly. In the layers, we want to uncover the low-dimensional $Z$-structure in a way that does not disregard information about predicting the output $Y$.

Principal component analysis (PCA), reduced rank regression (RRR), linear discriminant analysis (LDA), projection pursuit regression (PPR), and logistic regression are all shallow learners. For example, PCA reduces $X$ to $f_2(X)$ using a
singular value decomposition of the form
\begin{equation}
Z =  f_2(X) = W^\top X + b\,,\label{PCA_eq}
\end{equation}
where the columns of the weight matrix $W$ form an orthogonal basis for directions of greatest variance (which is in effect an eigenvector problem). Similarly, for the case of $X=(x_1,\ldots,x_p)$, PPR reduces $X$ to $f_2(X)$ by setting
$$
Z=f_2(X) = \sum^{N_1}_{i=1}g_i(W_{i1}x_1 + \ldots + W_{ip}x_p)\,.
$$
As stated before, these types of dimension reduction is
independent of $y$ and can easily discard information that is valuable for
predicting the desired output. Sliced inverse regression (SIR)~\cite{li1991sliced} overcomes this drawback somewhat by
estimating the layer function $f_2$ using data on both, $Y$ and $X$, but still operates independently of $f_1$.

Deep learning overcomes many classic drawbacks by \textit{jointly}
estimating $f_1$ and $f_2$ based on the full training data $\hat{X}=\{Y_i,X_i\}^T_{i=1}$, using information on $Y$ and $X$ as well as their relationships, and by using $L>2$ layers.
If we choose to use nonlinear layers, we can view a deep learning routine as a hierarchical nonlinear factor model or, more specifically, as a generalised linear model (GLM) with recursively defined nonlinear link functions.

\cite{diaconis_nonlinear_1984} use  nonlinear functions of linear combinations. The hidden factors $z_i= w_{i1}b_1 + \ldots + w_{ip}b_p$ represents a data reduction of the output matrix $X$. The model selection problem is to choose $N_1$ (how many hidden units).

\subsection{GANs: Generative Adversarial Networks}
GAN has two components  – Generator Neural Network and Discriminator Neural Network. The Generator Network $G:z\rightarrow x$ maps random $z$ to a sample from the target distribution $p(x)$ and the Discriminator Network $D(x)$ is a binary classifier with two classes: generated sample and true sample. 

We train GAN iteratively by switching between  generator and discriminator. This can be represented mathematically as
\begin{align*}
&\min_{\theta_G}\max_{\theta_D} V(D(x\mid \theta_D),G(z\mid \theta_G))\\
V(D,G) & = \mathrm{E}_{x\sim p(x)}\left[\log D(x)\right] + \mathrm{E}_{z\sim p(z)}\left[\log(1 - D(G(z))\right]
\end{align*}

In $V(D, G)$, the first term is a deviance that penalizes for misclassification of samples, the goal is to have it close to 1.  The second term is entropy that the data from random input (p(z)) passes through the generator, which then generates a fake sample which is then passed through the discriminator to identify the fakeness (aka worst case scenario). In this term, the discriminator tries to maximize it to 0 (i.e. the log probability that the generated data is fake is equal to 0). So overall, the discriminator is trying to maximize our function $V$.

On the other hand, the task of generator is exactly opposite, i.e. it tries to minimize the function $V$ so that the differentiation between real and fake data is a bare minimum. This, in other words is a cat and mouse game between generator and discriminator!

\section{Algorithmic Issues} \label{sec:algo}
\subsection{Training}
Let the training dataset be denoted by $ \hat{X} = \{ Y_i , X_i\}_{i=1}^T $. Once the activation functions, size and depth of the learner have been chosen, we need to solve the training problem of finding $(\hat{W} , \hat{b}) $ 
where $\hat{W}=(\hat{W}_1,\ldots,\hat{W}_L)$ and $\hat{b}=(\hat{b}_1,\ldots,\hat{b}_L)$ denote the learning parameters which we compute during training.
A more challenging problem, is training the size and depth
 $N_l , L $, which is known as the model selection problem. 
To do this, we need a training dataset $ \hat{X} = \{ Y_i , X_i\}_{i=1}^T $ of input-output pairs, a loss function $l(Y,\hat{Y})$ at the level of the output signal.
In its simplest form, we simply solve
\begin{equation}
{\rm arg\,min}_{W,b}\frac{1}{N} \sum_{i=1}^T l( Y_i , \hat{Y}( X_i)) \,,\label{Training_Eq1}
\end{equation}
An $L_2$-norm for a traditional least squares problem becomes a suitable error measure, and if we then minimise the loss function $l( Y_i , \hat{Y}( X_i)) = \|Y_i - \hat{Y}( X_i)\|^2_2\,,$
our target function \eqref{Training_Eq1} becomes the mean-squared error (MSE). 

It is common to add a regularisation penalty $ \phi(W,b)$ to avoid over-fitting and to stabilise our predictive rule. We combine this with the loss function
with a global parameter $ \lambda$ that gauges the overall level of regularisation.
We  now need to solve
\begin{equation}
{\rm arg\,min}_{W,b}\frac{1}{N} \sum_{i=1}^T l( Y_i , \hat{Y}( X_i)) + \lambda \phi(W,b)\,,\label{Training_Eq2}
\end{equation}
Again we compute a nonlinear predictor $\hat{Y}=\hat{Y}_{\hat{W},\hat{b}} (X)$ of the output $Y$ given the input $X$--the goal of deep learning.

In a traditional probabilistic model $p(Y| \hat{Y}(X))$ that generates the output $Y$ given the predictor $\hat{Y}(X)$, we have the natural loss
function $l(Y, \hat{Y} ) = - \log p( Y| \hat{Y} )$ as the negative log-likelihood.
For example, when predicting the probability of default, we have a multinomial logistic regression model which leads to a cross-entropy loss function. For multivariate normal models, which includes many financial time series, the $L_2$-norm of traditional least squares becomes a suitable error measure.

The common numerical approach for the solution of \eqref{Training_Eq2} is a form of stochastic gradient descent, which adapted to a deep learning setting is usually called \emph{backpropagation}~\cite{rumelhart1986learning}. One caveat of backpropagation in this context is the multi-modality of the system to be solved and the resulting slow convergence properties, which is the main reason why deep learning methods heavily rely on the availablility of large computational power. 

To allow for \emph{cross validation}~\cite{hastie_elements_2016} during training, we may split our training data into disjoint time periods of identical length, which is particularly desirable in financial applications where reliable time consistent predictors are hard to come by and have to be trained and tested extensively. Cross validation also provides a tool to decide  what levels of regularisation lead to good generalisation (i.e., predictive) rules, which is the classic variance-bias trade-off. A key advantage of cross validation, over traditional statistical metrics such as $t$-ratios and $p$-values, is
that we can also use it to assess the size and depth of the hidden layers, that is, solve the model selection problem of choosing $ N_l $ for $1 \leq l \leq L $ and $L$ using the same
predictive MSE logic. This ability to seamlessly solve the model selection and estimation problems is one of the reasons for the current widespread use of machine learning methods.


\subsubsection{Approximate Inference}
The recent successful approaches to develop efficient Bayesian inference algorithms for deep learning networks are based on the reparameterization techniques for calculating Monte Carlo gradients while performing  variational inference. Given the data $D = (X,Y)$, the variation inference relies on approximating the posterior $p(\theta \mid D)$ with a variation distribution  $q(\theta \mid D,\phi)$, where $\theta = (W,b)$. Then $q$ is found by minimizing the based on the Kullback-Leibler divergence between the approximate distribution and the posterior, namely
\[
\text{KL}(q \mid\mid p) = \int q(\theta \mid D, \phi)\log \dfrac{q(\theta \mid D, \phi)}{p(\theta\mid D)}d\theta.
\]
Since $p(\theta\mid D)$ is not necessarily tractable, we replace minimization of $\text{KL}(q \mid\mid p) $ with maximization of  evidence lower bound (ELBO)
\[
\text{ELBO}(\phi) = \int q(\theta \mid D,\phi)\log \dfrac{p(Y\mid X,\theta)p(\theta)}{q(\theta \mid D, \phi)}d\theta
\]
The $log$ of the total probability (evidence) is then
\[
\log p(D) =  \text{ELBO}(\phi) + \text{KL}(q \mid\mid p)
\]
The sum does not depend on $\phi$, thus minimizing $\text{KL}(q \mid\mid p)$ is the same that maximizing $\text{ELBO}(q) $. Also, since $\text{KL}(q \mid\mid p) \ge 0$, which follows from Jensen's inequality, we have $\log p(D) \ge  \text{ELBO}(\phi)$. Thus, the  evidence lower bound name.  The resulting maximization problem $\text{ELBO}(\phi) \rightarrow \max_{\phi}$ is solved using stochastic gradient descent. 

To calculate the gradient, it is convenient to write the ELBO as
\[
\text{ELBO}(\phi) = \int q(\theta \mid D, \phi)\log p(Y\mid X,\theta)d\theta - \int q(\theta \mid D,\phi) \log \dfrac{q(\theta\mid D, \phi)}{p(\theta)}d\theta
\]
The gradient of the first term $\nabla_{\phi}\int q(\theta \mid D, \phi)\log p(Y\mid X,\theta)d\theta = \nabla_{\phi}E_q\log p(Y\mid X,\theta)$ is not an expectation and thus cannot be calculated using Monte Carlo methods. The idea is to represent the gradient $\nabla_{\phi} E_q\log p(Y\mid X,\theta)$ as an expectation of some random variable, so that Monte Carlo techniques can be used to calculate it. There are two standard methods to do it. First, the log-derivative trick, uses the following identity $\nabla_x f(x) = f(x) \nabla_x \log f(x)$ to obtain $\nabla_{\phi} E_q\log p(Y\mid \theta)$.
Thus, if we select $q(\theta \mid\phi)$ so that it is easy to compute its derivative and generate samples from it, the gradient can be efficiently calculated using Monte Carlo methods.  Second, we can use the reparametrization trick by representing $\theta$ as a value of a deterministic function,  $\theta = g(\epsilon,x,\phi)$, where $\epsilon \sim r(\epsilon)$ does not depend on $\phi$. The derivative is given by
\begin{align*}
\nabla_{\phi} E_q\log p(Y\mid X, \theta) &= \int r(\epsilon)\nabla_{\phi}\log p(Y\mid  X, g(\epsilon,x,\phi))d \epsilon\\
& = E_{\epsilon} [\nabla_{g}\log p(Y\mid X, g(\epsilon,x,\phi))\nabla_{\phi}g(\epsilon,x,\phi)].
\end{align*}
The reparametrization is trivial when $q(\theta \mid D,\phi) = N(\theta \mid \mu(D,\phi), \Sigma(D,\phi))$, and $\theta = \mu(D,\phi)  + \epsilon\Sigma(D,\phi),~\epsilon\sim N(0,I)$. \cite{kingma2013auto}  propose using $\Sigma(D,\phi) = I$ and representing $\mu(D,\phi)$ and $\epsilon$ as outputs of a neural network (multi-layer perceptron), the resulting approach was called variational auto-encoder. A generalized reparametrization has been proposed by \cite{ruiz2016generalized} and combines both log-derivative and reparametrization techniques by  assuming that $\epsilon$ can depend on $\phi$. 


\subsection{Dropout}
To avoid overfitting in the training process, \emph{dropout} is the technique~\cite{srivastava_dropout:_2014} of removing input dimensions in $X$ randomly with probability $p$. In effect, this replaces the input $X$ by $ D \star X $, where $ \star$ denotes the element-wise product
and $D$ is a matrix of Bernoulli $\mathcal{B}(p)$ random variables. For example, setting $l(Y,\hat{Y})=\|Y-\hat{Y}\|^2_2$ (to minimise the MSE as explained above) and $\lambda=1$, marginalised over the randomness, we then have a new objective
$$
{\rm arg \; min}_W \; \mathbb{E}_{ D \sim {\rm Ber} (p) } \Vert Y - W ( D \star X ) \Vert^2_2\,,
$$
which is equivalent to
$$
{\rm arg \; min}_W \;  \Vert Y - p W X \Vert^2_2 + p(1-p) \Vert \Gamma W \Vert^2_2\,,
$$
where $ \Gamma = ( {\rm diag} ( X^\top X) )^{\frac{1}{2}} $. We can also interpret this last expression as a Bayesian ridge regression with a $g$-prior~\cite{wager2013dropout}. Put simply, dropout reduces the likelihood of over-reliance on small sets of input data in training. 

\begin{equation}
{\rm arg\,min}_{W,b}\frac{1}{N_l} \sum_{i=1}^T l(Y_i , \hat{Y_i}) + \lambda \phi(W,b),\label{Training_Eq}
\end{equation}

Another application of dropout regularisation is the choice of the number of hidden units in a layer (if we drop units of the hidden rather than the input layer and then establish which probability $p$ gives the best results). It is worth recalling though, as we have stated before, that one of the dimension reduction properties of a network structure
is that once a variable from a layer is dropped, all terms above it in the network also disappear. This is just the nature of a composite structure for the deep predictor in \eqref{DLComp}.

\subsection{Batch Normalization}
Dropout is mostly a technique for regularization. It introduces noise into a neural network to force the neural network to learn to generalize well enough to deal with noise. 

Batch normalization \cite{ioffe2015batch} is mostly a technique for improving optimization. As a side effect, batch normalization happens to introduce some noise into the network, so it can regularize the model a little bit.

We normalize the input layer by adjusting and scaling the activations. For example, when we have features from 0 to 1 and some from 1 to 1000, we should normalize them to speed up learning. If the input layer is benefiting from it, why not do the same thing also for the values in the hidden layers, that are changing all the time, and get 10 times or more improvement in the training speed.

Batch normalization reduces the amount by what the hidden unit values shift around (covariance shift). If an algorithm learned some $X$ to $Y $mapping, and if the distribution of $X$ changes, then we might need to retrain the learning algorithm by trying to align the distribution of $X$ with the distribution of $Y$.  Also, batch normalization allows each layer of a network to learn by itself a little bit more independently of other layers. When you have a large dataset, it's important to optimize well, and not as important to regularize well, so batch normalization is more important for large datasets. You can of course use both batch normalization and dropout at the same time

We can use higher learning rates because batch normalization makes sure that there no extremely high or low activations. And by that, things that previously couldn't get to train, it will start to train. It reduces overfitting because it has a slight regularization effects. Similar to dropout, it adds some noise to each hidden layers' activations. Therefore, if we use batch normalization, we will use less dropout, which is a good thing because we are not going to lose a lot of information. However, we should not depend only on batch normalization for regularization; we should better use it together with dropout.
How does batch normalization work? To increase the stability of a neural network, batch normalization normalizes the output of a previous activation layer by subtracting the batch mean and dividing by the batch standard deviation.

However, after this shift/scale of activation outputs by some randomly initialized parameters, the weights in the next layer are no longer optimal. SGD ( Stochastic gradient descent) undoes this normalization.
Consequently, batch normalization adds two trainable parameters to each layer, so the normalized output is multiplied by a ``standard deviation'' parameter (gamma) and adds a ``mean'' parameter (beta). In other words, batch normalization lets SGD do the denormalization by changing only these two weights for each activation, as follows:
\begin{align*}
\mu_B & = \dfrac{1}{m}\sum_{i=1}^{m}x_i,~~
\sigma_B^2 = \dfrac{1}{m}\sum_{i=1}^{m}(x_i-\mu_B)^2\\
\hat{x}_i &= \dfrac{x_i - \mu_B}{\sqrt{\sigma_B^2 + \epsilon}},~~
y_i   = \gamma \hat{x}_i + \beta = \mathrm{BN}_{\gamma,\beta}(x_i).
\end{align*}

\section{Deep Learning Theory} \label{sec:gp}
There are two questions for which we  do not yet have a comprehensive and a satisfactory answer. The first is, how to choose a deep learning architecture for a given problem. The second is why the deep learning model does so well on out-of-sample data, i.e. generalize.

To choose an appropriate architecture, from practical point of view, techniques such as Dropout or universal architectures~\cite{kaiser2017one} allow us to spend less time on choosing an architecture. Also some recent Bayesian theory sheds a light on  the problem\\\cite{polson2018posterior}. However, it is still required to go through a trial-and-error process and empirically evaluate large number of models, before an appropriate architecture could be found. On the other hand, there are some theoretical results that shed a light on the architecture choice process. 

It was long well known that shallow networks are universal approximators and thus can be used to learn any input-output relations. The first result in this direction was obtained by Kolmogorov~\cite{kolmogorov57} who has shown that any multivariate function can be exactly  represented using operations of addition and superposition on univariate functions. Formally, there exist continuous functions $\psi^{pq}$, defined on $[0,1]$ such that each continuous real function $f$ defined on $[0,1]^n$ is represented as
\[
g(x_1,\ldots,x_n) = \sum_{q=1}^{2n+1}\chi_q\left(\sum_{p=1}^{n}\psi^{pq}(x_p)\right),
\]
where each $\chi_q$ is a continuous real function. This representation is a generalization of earlier results~\cite{kolmogorov56,arnold1963}. In~\cite{kolmogorov56} it was shown that every continuous multivariate function can be represented in the form of a finite superposition  of continuous functions of not more than three variables. Later Arnold used that result to solve Hilbert's thirteenth problem~\cite{arnold1963}.

However, results of Kolmogorov and Arnold do not have practical application. Their proofs are not constructive and do not demonstrate how functions  $\chi$ and $\psi$ can be computed. Further, \cite{girosi1989representation} and references therein show that those functions are not smooth, while in practice smooth functions are used. Further, while functions $\psi$ form a universal basis  and do not depend on $g$, the function $\chi$ does depend on the specific form of function $g$. More practical results were obtained by Cybenko~\cite{cybenko1989approximation} who showed that a shallow network with sigmoidal activation function can arbitrarily well approximate any continuous function on the $n$-dimensional cube $[0,1]^n$. Generalization for a broader class of activation functions was derived in \cite{hornik1991approximation}.

Recently, it was shown that deep architectures require smaller number of parameters compared to shallow ones to approximate the same function and thus more computationally viable. It was observed empirically that deep learners perform better. \cite{montufar2014number} provide some geometric intuition and show that the number of open box regions generated by deep network is much lager than those generated by a shallow one with the same number of parameters. \cite{telgarsky2015representation} and \cite{safran2016depth} provided specific examples of classes of functions that require an exponential number of parameters as a function of input dimensionality to be approximated by shallow networks.

\cite{poggio2017and} provides a good overview of results on complexity of shallow and deep networks required to approximate different classes of functions. For example, for  shallow (one-layer) network
\[
g(x)\approx F(x) = \sum_{i=i}^{N}a_kf(w_k^Tx + b_k)
\]
with $f$ being infinitively differentiable and not a polynomial, it is required  that $N = \mathcal{O}(\epsilon^{-n/m})$. Here $x\in [0,1]^n$ (n-dimensional cube), $g(x)$ is is differentiable up to order $m$ and $\epsilon$ is the required accuracy of the approximation, e.g. $\max_x|g(x) - f(x)| < \epsilon$.  Thus, the size of neurons required is exponential in the number of input dimensions, i.e. the curse of dimensionality. 

Meanwhile, if function $g(x)$ has a special structure, then deep learner avoids the curse of dimensionality problem. 

Specifically, let $g(x): R^n \rightarrow R$ be a $\mathcal{G}$--function, which is defined as follows. Source nodes are components of the input $x_1,\ldots,x_n$, and there is only one sink node (value of the function. Each node $v$ between the source and sink is a local function of dimensionality $d_v << n$ of low dimensionality and $m_v$ times differentiable.  
For example, in $\mathcal{G}$-function $f(x_1, \cdots, x_8) = h_3(h_{21}(h_{11} (x_1, x_2), h_{12}(x_3,
x_4)), h_{22}(h_{13}(x_5, x_6), h_{14}(x_7, x_8)))$ each  $h$ is ``local'', e.g. requires only two-dimensional input to be computed, it has two hidden nodes and each  has two inputs ($d_v=2$ for every hidden node). The required complexity  of deep network to represent such a function is given by
\[
N_s = \mathcal{O}\left(\sum_{v \in V} \epsilon ^{-d_v/m_v}\right).
\]

Thus, when the target function has local structures as defined for a $\mathcal{G}$--function, deep neural networks allow us to avoid curse of dimensionality. 

In~\cite{lin2017does} the authors similarly show how the specific structure of the target function to be approximated can be exploited to avoid the curse of dimensionality. Specifically, properties frequently encountered in physics, such as  symmetry, locality, compositionality, and polynomial log-probability translate into exceptionally simple neural networks and are shown to lead to low complexity deep network approximators.  

The second question of why deep learning models do not overfit and generalize well to out-of-sample data has received less attention in the literature. It was shown in \cite{zhang2016understanding} that regularization techniques do not explain a surprisingly good out-of-sample performance. Using a series of empirical experiments it was shown that deep learning models can learn white noise. A contradictory result was obtained in~\cite{liao2018classical} and shows that, for a DL model with exponential loss function and appropriately normalized, there is a linear dependence of test loss on training loss.

\bibliography{ref}
\bibliographystyle{apalike}

\end{document}